\documentclass[]{itatnew}
\usepackage{footnote}
\usepackage{tablefootnote}
\usepackage{tabularx}
\usepackage{multirow}
\usepackage{booktabs}
\usepackage{subcaption}
\makesavenoteenv{tabular}
\usepackage[square,sort,comma,numbers]{natbib}
\usepackage{graphicx}
\usepackage[utf8]{vietnam}
\usepackage[english]{babel}
\usepackage{natbib}
\usepackage{todonotes}
\usepackage{pgfplots}
\usepackage{makecell}
\usepackage{paralist}

\begin{document}
\title{Syntax Representation in Word Embeddings and Neural Networks -- A Survey}

\author{Tomasz Limisiewicz and David Mare\v{c}ek}

\institute{Institute of Formal and Applied Linguistics, Faculty of Mathematics and Physics, Charles University\\
\email{\{limisiewicz,marecek\}@ufal.mff.cuni.cz}}

\maketitle

\begin{abstract}

Neural networks trained on natural language processing tasks capture syntax even though it is not provided as a supervision signal.
This
indicates that syntactic analysis is essential to the understating of language in artificial intelligence systems.
This overview paper covers approaches of evaluating the amount of syntactic information
included in the representations of words for different neural network architectures.
We mainly summarize research on English monolingual data on language modeling tasks and multilingual data for neural machine translation systems and multilingual language models.  
We describe which pre-trained models and representations of language are best suited for transfer to syntactic tasks.



\end{abstract}

\section{Introduction}

Modern methods of natural language processing (NLP) are based on complex neural network architectures, where language units are represented in a metric space \cite{mikolov2013efficient, pennington2014glove, peters2018deep, devlin2019bert, radford2019language}. Such a phenomenon allows us to express linguistic features (i.e., morphological, lexical, syntactic) mathematically. 

The method of obtaining such representation and their interpretations were described in multiple overview works. Almeida and Xex\'eo surveyed different types of static word embeddings \cite{almeida2019word}, and Liu et al. \cite{liu2020contextual} focused on contextual representations found in the most recent neural models. Belinkov and Glass \cite{belinkov2017evaluating} surveyed the strategies of interpreting latent representation. Best to our knowledge, we are the first to focus on the syntactic and morphological abilities of the word representations. We also cover the latest approaches, which go beyond the interpretation of latent vectors and analyze the attentions present in state-of-the-art Transformer models. 


\section{Vector Representations of Words}
\label{sec:model-types}

This section introduces several types of architectures that we will analyze in this work.

\subsection{Static Word Embeddings}

In the classical methods of language representation, each word is assigned a vector regardless of its current context. In the Latent Semantic 
Analysis \cite{deerwester1990indexingby}, the representation was obtained by counting word frequency across documents on distinct subjects.

In more recent approaches, a shallow neural network is used to predict each word based on context (Word2Vec \cite{mikolov2013efficient}) or approximate the frequency of coocurence for a pair of words (GloVe \cite{pennington2014glove}). One explanation of the effectiveness of these algorithms is the distributional hypothesis \cite{harris54}: "words that occur in the same contexts tend to have similar meanings".

\subsection{Contextual Word Vectors in Recurrent Networks}

The main disadvantage of the static word embeddings is that they do not take into account the context of words. This is especially an issue for languages rich in words that have multiple meanings.

The contextual embeddings introduced in \cite{peters2018deep} and \cite{mccann2017learned}
are able to encode both words and their contexts.
They are based on recurrent neural networks (RNN) and are typically trained on language modeling or machine translation tasks using large text corpora.
The outputs of the RNN layers are context-dependent representations that are proven to perform well when used as inputs for other NLP tasks with much less training data available.


Another improvement of context modeling was possible thanks to the attention mechanism \cite{bahdanau2014neural}. It allowed passing the information from the most relevant part of the RNN encoder, instead of using only the contextual representation of the last token.

\subsection{Contextual Representation in Transformers}

\begin{figure}[t]
    \centering
    \includegraphics[width=\linewidth]{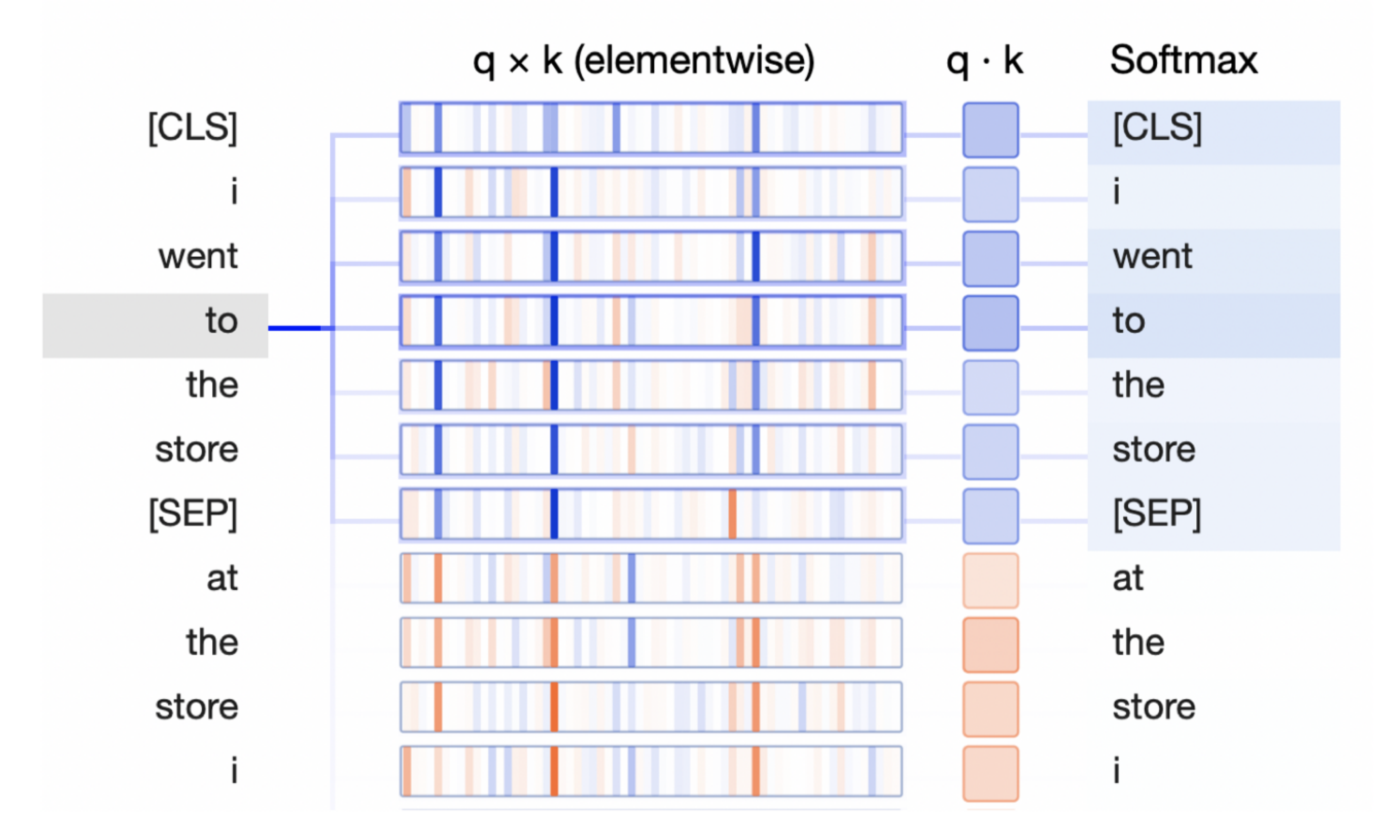}
    \caption{Visualization of attention mechanism in Transformer architecture. It shows which parts of the text are important to compute the representation for the word “to”. Created in BertViz framework \cite{vig2019transformervis}.}
    \label{fig:attn-visu}
\end{figure}

The most recent and widely used architecture is the Transformer \cite{vaswani2017attention}. It consists of several (6 to 24) layers, and each token position in each layer has the ability to attend to any position in the previous layer using a self-attention mechanism.   
Training such architecture can be easily parallelized since individual tokens can be processed independently; their positions are encoded within the input embeddings.
An example of visualization of attention distribution computed in Transformer trained for language modeling (BERT \cite{devlin2019bert}) is presented in Figure~\ref{fig:attn-visu}. 

In addition to vectors, Transformer includes latent representation in the form of self-attention weights, which are two-dimensional matrices. We summarize the research on the syntactic properties of attention weights in Section~\ref{sec:matrices-observation}. 



    


\section{Measures of Syntactic Information}
\label{sec:syntactic-measures}

This sections describes the metrics used to evaluate syntactic information captured by the word embeddings and latent representation.

\subsection{Syntactic Analogies}

\begin{figure}[]
    \centering
    \includegraphics[width=0.8\linewidth]{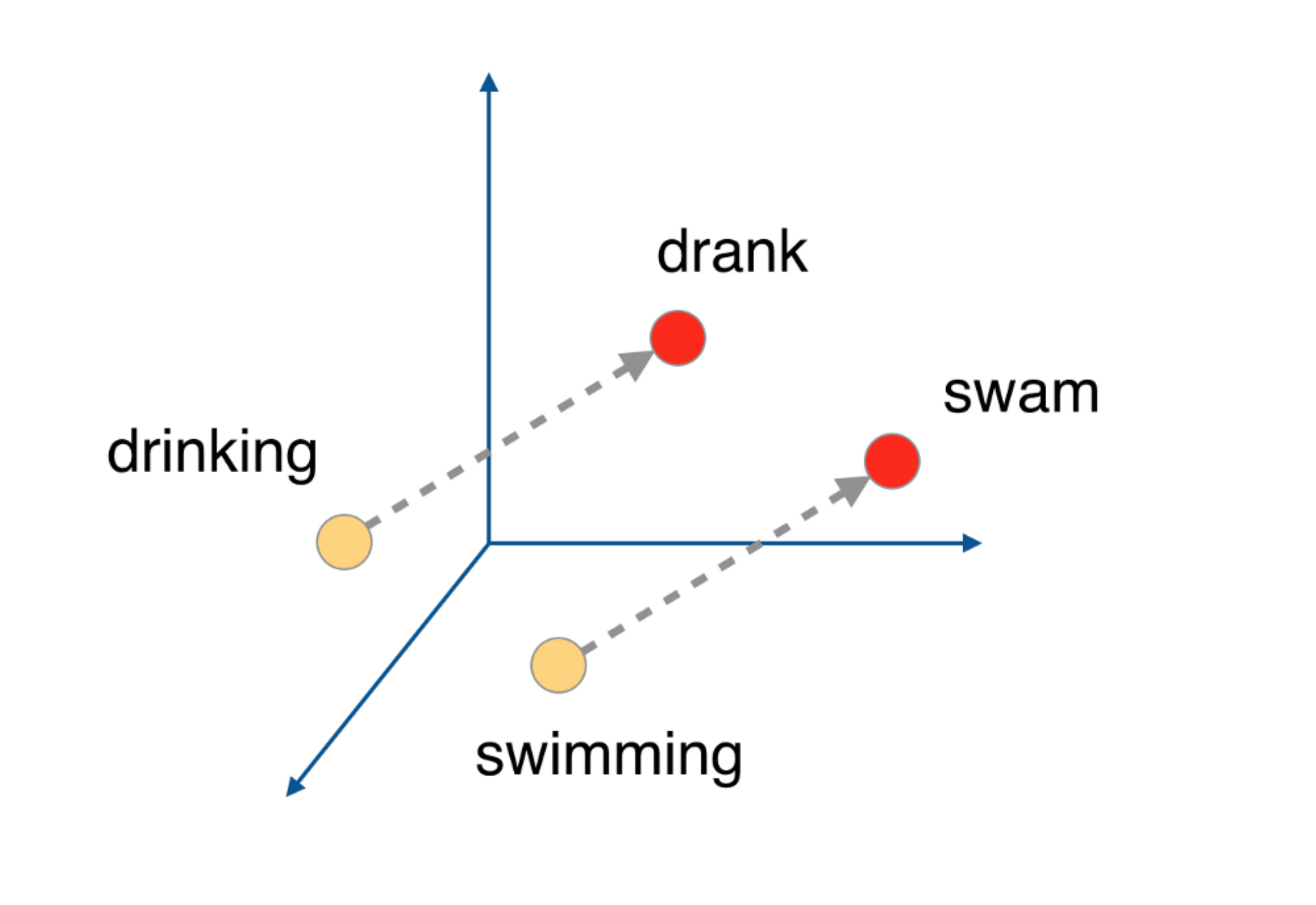}
    \caption{Spatial distribution of word embeddings depends on syntactic roles of words (visualization created by Ashutosh Singh).}
    \label{fig:static-we}
\end{figure}

In the recent revival of word embeddings\cite{mikolov2013efficient, pennington2014glove}, a strong focus was put on examining the phenomenon of encoding analogies in multidimensional space.
That is to say, the shift vector between pairs of analogous words is approximately constant, e.g., the pairs \emph{drinking -- drank}, \emph{swimming -- swam} in Figure~\ref{fig:static-we}. 

Syntactic analogies of this type are particularly relevant for this overview. They include the following relations:
adjective -- adverb; singular -- plural; adjective -- comparative -- superlative; verb -- present participle -- past participle. The syntactic  analogy is usually evaluated on Google Analogy Test Set \cite{mikolov2013efficient}. \footnote{The test set is called syntactic by authors; nevertheless, it mostly focuses on morphological features.}

An evaluation example consists of two word pairs represented by the embeddings: $(v_1, v_2), (u_1, u_2)$. We compute the analogy shift vector as the difference between embeddings of the first pair $s = v_2 - v_1$. The result is positive if the nearest word embedding to the vector $u_1 + s$ is $u_2$.
\begin{equation}
WA = \frac{|\{(v_1, v_2, u_1, u_2): u_2 \approx u_1 + v_2 - v_1\}|}{|\{(v_1, v_2, u_1, u_2)\}|}
\end{equation}

\subsection{Sequence Tagging}

Sequence tagging is a multiclass classification problem. The aim is to predict the correct tag for each token of a sequence. A typical example is the part of speech (POS) tagging. The accuracy evaluation is straightforward: the number of correctly assigned tags is divided by the number of tokens. 



\subsection{Syntactic structure prediction}


The inference of reasonable syntactic structures from word representations is the most
challenging
task covered in our survey. There are attempts to predict both the dependency\cite{hewitt-manning-2019-structural, raganatotiedemann2018analysis, limisiewicz2020universal, clark2019does} and constituency trees \cite{marecekrosa2019balustrades, kim_are_2020}.
Dependency trees are evaluated using unlabeled attachment score (UAS) or its undirected variant (UUAS):
\begin{equation}
UAS = \frac{\#correctly\_attached\_words}{\#all\_words}
\end{equation}
The equation for Labeled Attachment Score is the same, but it requires predicting a dependency label for each edge. For constituency, trees we define precision (P) and recall (R) for correctly predicted phrases.
\begin{equation}
P = \frac{\#correct\_phrases}{\#gold\_phrases},\quad
R = \frac{\#correct\_phrases}{\#predicted\_phrases}
\end{equation}

Usually, $F1$ is reported, which is a harmonic mean of precision and recall.


\subsection{Attention's Dependency Alignment}

In Section~\ref{sec:matrices-observation} we describe the examination of syntactic properties of self-attention matrices. It can be evaluated using \emph{Dependency Alignment} \cite{vigbelinkov2019analyzing} which sums the attention weights at the positions corresponding to the pairs of tokens forming a dependency edge in the tree.
\begin{equation}
DepAl_{A} =\frac{\sum_{(i,j) \in E} A_{i,j}}{\sum_{i=1}^N\sum_{j=1}^N  A_{i,j}}
\end{equation}

\emph{Dependency Accuracy} \cite{voita2019analyzing, clark2019does, limisiewicz2020universal} is an alternative metric; for each dependency label it measures how often the relation's governor/dependent is the most attended token by the dependent/governor. 
\begin{equation}
DepAcc_{l,d, A} =\frac{|\{(i,j) \in E_{l,d} : j = \arg\max A_{i,\cdot} \}|}{|E_{l,d}|}
\end{equation}
\textbf{Notation:} $E$ is a set of all dependency tree edges and $E_{l,d}$ is a subset of the edges with the label $l$ and with direction $d$, i.e., in dependent to governor direction the first element of the tuple $i$ is dependent of the relation and the second element $j$ is the governor; $A$ is a self-attention matrix and $A_{i, \cdot}$ denotes $i^{th}$ row of the matrix; $N$ is the sequence length.

\section{Morphology and Syntax in Word Embeddings and Latent Vectors}

In this section, we summarize the research on the syntactic information captured by vector representations of words. We devote a significant attention to POS tagging, which is a popular evaluation objective. Even though it is a morphological task, it is highly relevant to syntactic analysis.

\label{sec:vectors-observation}


\subsection{Syntactic Analogies}

The first wave of research on the vector representation of words focused on the statistical distribution of words across distinct topics -- Latent Semantic Analysis \cite{deerwester1990indexingby}. It captured statistical properties of words, yet
there were 
no positive results in syntactic analogies retrieval nor encoding syntax. 

Google Analogy Test Set was released together with a popular word embedding algorithm Word2Vec~\cite{mikolov2013efficient}. One of the exceptional properties of this method was its high accuracy in the analogy tasks. In particular, the best configuration found the correct syntactic analogy in 68.9 \% of cases. 

The GloVe embeddings improved the results on syntactic analogies to 69.3\% \cite{pennington2014glove}.
Much more significant improvement was reported for semantic analogies. They also outperform the variety of other vectorization methods.

In \cite{mikolov2013linguistic} a simple recurrent neural network was trained by language modeling objective. The word representation is taken from the input layer. 
The evaluation from \cite{mikolov2013efficient} shows that Word2Vec performs better in syntactic analogy task. This observation is surprising because representations from RNN were proven effective in transfer to other syntactic tasks (we elaborate on that in Sections~\ref{sec:POS-tagging} and~\ref{sec:structure-induction}).  We think that possible explanations could be: 1. the techniques of RNN training have crucially improved in recent years; 2. syntactic analogy focuses on particular words, while for other syntactic tasks, the context is more important.

\begin{figure*}[t]
    \centering
    \begin{subfigure}{0.46\linewidth}
        \begin{center}
        \resizebox{\linewidth}{!}{\input{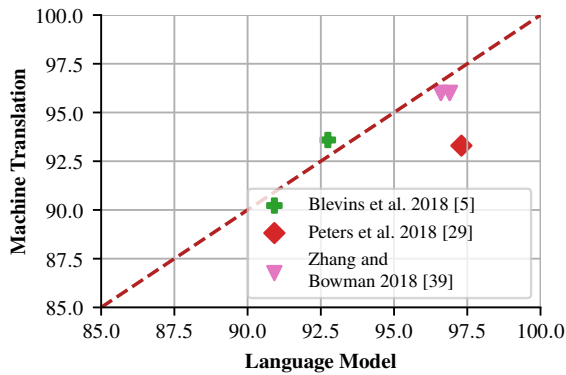}}
        \end{center}
        \caption{Neural machine translation compared with language modeling pre-training} 
        \end{subfigure}\hfill
        \begin{subfigure}{0.46\linewidth}

        \begin{center}
        \resizebox{\linewidth}{!}{\input{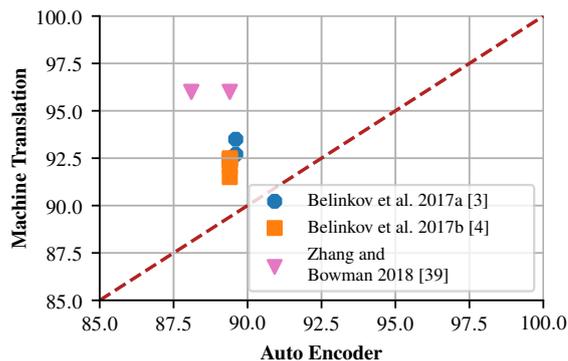}}
        \end{center}
        \centering
        \caption{Neural machine translation compared with auto encoder pre-training}
    \end{subfigure}
    \caption{Accuracy of POS tag probing from RNN representation by the pre-training objective.}
    \label{fig:POS-lm-nmt-ae}
\end{figure*}

\begin{figure}[h]
    \centering

    \resizebox{0.9\linewidth}{!}{\input{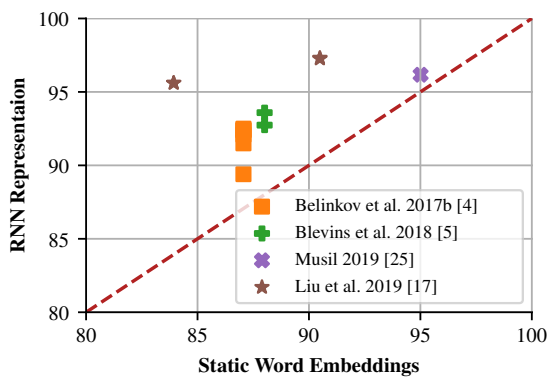}}

    \caption{Accuracy of POS tag probing from RNN latent vectors compared with static word embeddings}
    \label{fig:POS-rnn-swe}
\end{figure}

\subsection{Part of Speech Tagging}
\label{sec:POS-tagging}

%
Measuring to what extent a linguistic feature such as POS is captured in word representations is
usually performed by the method called \emph{probing}. In probing, the parameters of the pretrained network are fixed, the output word
representations
are computed as in the inference mode and then fed to a simple neural layer. Only this simple layer is optimized for a new task. 



The number of probing experiments rose with the advent of multilayer 
\footnote{Layer numbering in this work: We are numbering layers starting from one for the layer closest to the input. Please note that original papers may use different numbering.}
RNNs trained for language modeling and machine translation. 

Belinkov et al.~\cite{belinkov2017neural} probe a recurrent neural machine translation (NMT) system with four layers to predict part of speech tags (along with morphological features). They use Arabic, Hebrew, French, German, and Czech to English pairs. They observe that adding a character-based representation computed by a convolutional neural network in addition to word-embedding input is beneficial, especially for morphologically rich languages. 

In a subsequent study~\cite{belinkov2017evaluating}, the source language of translation now is English and the experiments are conducted solely for this language. It is noted that the most morphosyntactic representation is usually obtained in the middle layers of the network.


The influence of using a particular objective in pre-training RNN model is comprehensively analyzed by Blevins et al. \cite{blevins2018deep}. They pre-train models on four objectives: syntactic parsing, semantic role labeling, machine translation, and language modeling. The two former objectives may reveal morphosyntactic information to a larger extent than other mentioned here settings. Particularly, the probe of RNN syntactic parser achieves near-perfect accuracy in part of speech tagging.

The introduction of ELMo \cite{peters2018deep} brought a remarkable advancement in transfer learning from the RNN language model to a variety of other NLP tasks. The authors examined POS capabilities of the representations and compared the results with the neural machine translation system CoVe \cite{mccann2017learned}, which also uses RNN architecture.

Zhang et al.~\cite{zhang2018language} perform further experiments with CoVe and ELMo. They demonstrate that language modeling systems are better suited to capture morphology and syntax in the hidden states than machine translation, if comparable amounts of data are used to train both systems. Moreover, the corpora for language modeling are typically more extensive than for machine translation, which can further improve the results.

Another comprehensive evaluation of morphological and syntactic capabilities of language models was conducted by Liu et al.~\cite{liu2019linguistic}. Probing was applied to a language model based on the Transformer architecture (BERT) and compared with ELMo and static word embeddings (Word2Vec). They observe that the hidden states of Transformer do not demonstrate a major increase in probed POS accuracy over the RNN model, even though it is more complex and consists of a larger number of parameters.

POS tag probing was also performed for languages other than English. For instance, Musil~\cite{musil2019examining} trains translation systems (with RNN and Transformer architecture) from Czech to English and examines the learned input embeddings of the model and compares them to a Word2Vec model trained on Czech.



In Figures~\ref{fig:POS-lm-nmt-ae} and~\ref{fig:POS-rnn-swe}, we present a comparison of different settings for POS tag probing. Each point denotes a pair of results obtained in the same paper and the same dataset, but with different types of embeddings or pretraining objectives.
Therefore, we can observe that the setting plotted on the y-axis is better than the x-axis setting if the points are above identity function (red dashed line). We cannot say whether a method represented by another point performs better, as the evaluation settings differ.

Figure~\ref{fig:POS-rnn-swe} clearly shows that the RNN contextualization helps in part of speech tagging. As expected, the information about neighboring tokens is essential to predict morphosyntactic functions of words correctly. It is especially true for the homographs, which can have various part of speech in different places in the text.

The influence of RNN's pre-training task is presented in Figure~\ref{fig:POS-lm-nmt-ae}. Machine translation captures much better POS information than auto-encoders, which can be interpreted as translation from and to the same language. It is likely that the latter task is straightforward and therefore does not require to encode morphosyntax in the latent space. The difference between the results of machine translation and language modeling is small. Zhang et al.~\cite{zhang2018language} show that using a larger corpus for pre-training improves the POS accuracy. The main advantage of language models is that monolingual data is much easier to obtain than parallel sentences necessary to train a machine translation system.



\subsection{Syntactic Structure Induction}
\label{sec:structure-induction}

Extraction of dependency structure is more demanding because instead of prediction for single tokens, every pair of words need to be evaluated. 

Blevins et al.~\cite{blevins2018deep} propose a feed-forward layer on top of a frozen RNN representation to predict whether a dependency tree edge connects a pair of tokens. They concatenate the vector representation of each of the words and their element-wise product. Such a representation is fed as an input to the binary classifier. It only looks on a pair of tokens at a time, therefore predicted edges may not form a valid tree.


Another approach, induction of the whole syntactic structures from latent representations was proposed by Hewitt and Manning~\cite{hewitt-manning-2019-structural}. 
Their syntactic probing is based on training a matrix which is used to transform the output of network's layers (they use BERT and ELMo). The objective of the probing is to approximate dependency tree distances between tokens \footnote{Tree distance is the length of the tree path between two tokens} by the L2 norm of the difference of the transformed vectors. Probing produces the approximate syntactic pairwise distances for each pair of tokens. The minimum spanning tree algorithm is used on the distance matrix to find the undirected dependency tree. The best configuration employs the 15th layer of BERT large and induces treebank with 82.5\% UAS on Penn Treebank with Stanford Dependency annotation (relation directions and punctuation were disregarded in the experiments). The result for BERT is significantly higher than for ELMo, which gave 77.0\% when the first layer was probed.

The paper also describes an alternative method of approximating the syntactic depth by the L2 norm of latent vector multiplied by a trainable matrix. The estimated depths allow prediction of the root of a sentence with 90.1\% accuracy when representation from the 16th layer of BERT large is probed.

\subsection{Multilingual Representations}

The subsequent paper by Chi et al.~\cite{chi2020finding} applies the setting from \cite{hewitt-manning-2019-structural} to the multilingual language model mBERT. They train syntactic distance probes on 11 languages and compare UAS of induced trees in four scenarios: 1. training and evaluating on the same languages; 2. training on a single language, evaluating on a different one; 3. training on all languages except the evaluation one; 4. training on all languages, including the evaluation one. They demonstrate that the transfer is effective as the results in all the configurations outperform the baselines\footnote{There are two baselines: right-branching tree and probing on randomly initialized mBERT without pretraining}. Even in the hardest case -- zero-shot transfer from just one language, the result is at least 6.9 percent points above the baselines (for Chinese). Nevertheless, for all the languages, no transfer-learning setting can beat the training and evaluating a probe on the same language.

\begin{figure}[]
    \centering
    \includegraphics[width=\linewidth]{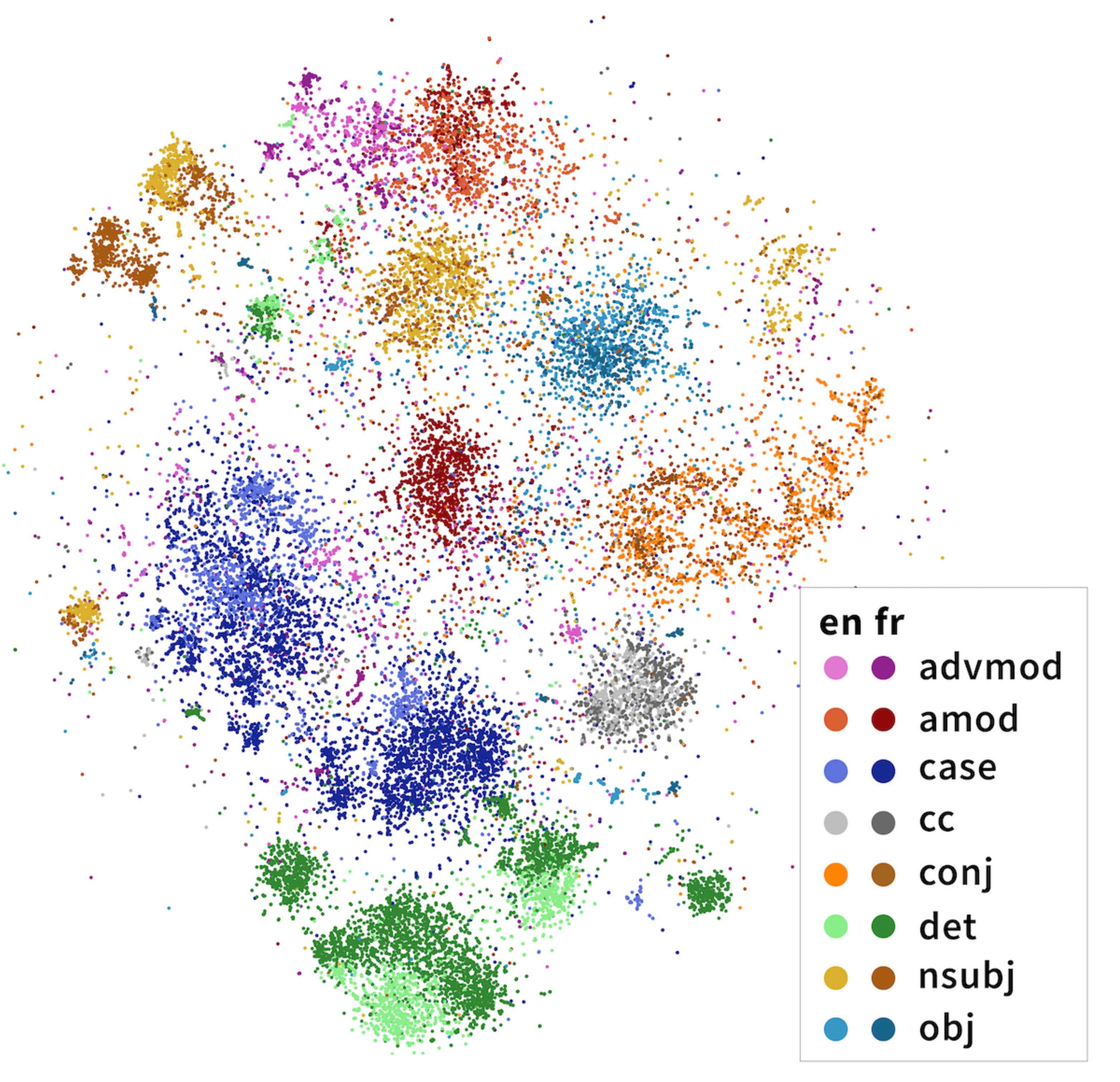}
    \caption{Two dimensional t-SNE visualization of probed mBERT embeddings from \cite{chi2020finding}. Analysis of the clusters shows that embeddings encode information about the type of dependency relations and, to a lesser extent, language.}
    \label{fig:mbert-probes}
\end{figure}

The paper includes analysis of intrinsic features of the BERT's vectors transformed by a probe. Noticeably,  the vector differences between the representations of words connected by dependency relation are clustered by relation labels, see figure~\ref{fig:mbert-probes}.

Multilingual BERT embeddings are also analyzed by Wang et al.~\cite{Wang-2019}. They show that even for the multilingual vectors, the results can be improved by projecting vector spaces across languages. They use Biaffine Graph-based Parser by Dozat and Manning~\cite{dozat2016deep}, which consists of multiple RNN layers. Therefore, the experiment is not strictly comparable with probing as the most of syntactic information is captured by the parser, and not by the embeddings. The article compares different types of vector representations fed as an input to the parser. It is demonstrated that cross-lingual transformation on mBERT embedding improves the results significantly in LAS of parser trained on English and evaluated on 14 languages (including English); on average, from 60.53\% to 63.54\%. In comparison to other cross-lingual representations, the proposed method outperforms transformed static embeddings (FastText with SVD) and also slightly outperforms contextual embeddings  (XLM).








\section{Syntax in Transformer's Attention Matrices}
\label{sec:matrices-observation}

Besides the vector representations of individual tokens, the Transformer architecture offers another representation with a possible syntactic interpretation – the weights of the self-attention heads. In each head, information can flow from each token to any other one.  These connections may be easily analyzed and compared to syntactic relations proposed by linguists. In this section, we will summarize different approaches of extracting syntax from attention. We present the methods both for dependency and constituency structures.


\subsection{Dependency Trees}

\begin{figure}[]
\centering
\input{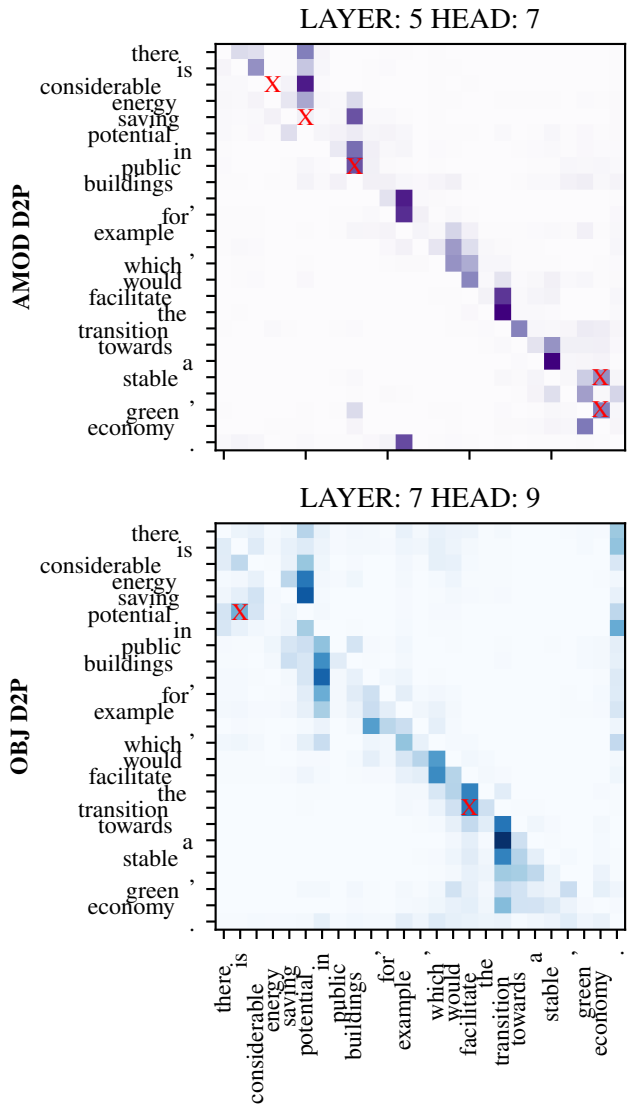}
\caption{Self-attentions in particular heads of a language model (BERT) aligns with dependency relation adjective modifiers and objects. The gold relations are marked with Xs.}
\label{fig:lm-heads}
\end{figure}

\begin{figure}[]
\centering
\input{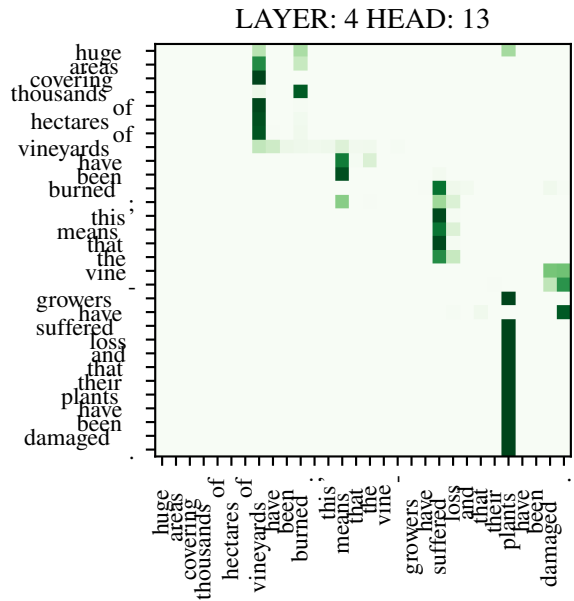}
\caption{Balustrades observed in NMT's encoder tend to overlap with syntactic phrases.}
\label{fig:balusters}
\end{figure}

\begin{savenotes}
\begin{table*}[]
\small
\centering
\begin{tabular}{@{}|p{3cm}|p{2.9cm}|p{1.8cm}|p{2.2cm}|p{3cm}|p{1.7cm}|@{}}\hline
 Research & Transformer Model &  Type of tree & Syntactic \newline evaluation & Evaluation data & Percentage of syntactic heads \\ \toprule
Raganato and \newline Tiedemann 2019 \cite{raganatotiedemann2018analysis} & NMT  Encoder \newline  (6 layers 8 heads) & Dependency &  Tree induction & PUD \cite{nivre2017ud}& 0\% - 8\%\footnotemark \\ \hline
Vig and Belinkov 2019 \cite{vigbelinkov2019analyzing} & LM (GPT-2) & Dependency  & Dependency Alignment & Wikipedia (automatically annotated) & --- \\ \hline
Clark et al. 2019 \cite{clark2019does} & LM (BERT) & Dependency  & Dependency \newline Accuracy, \newline Tree induction & WSJ Penn Treebank \cite{marcus1993penn}& ---  \\ \hline

Voita et al. 2019 \cite{voita2019analyzing} & NMT Encoder  \newline (6 layers 8 heads) & Dependency & Dependency \newline Accuracy & WMT, OpenSubtitles \cite{lison2018opensubtitles}  (both automatically annotated) & 15\% - 19\% \\ \hline

Limisiewicz et al. 2020 \cite{limisiewicz2020universal} & LMs \newline (BERT, mBERT) & Dependency  & Dependency \newline Accuracy, \newline Tree induction & PUD \cite{nivre2017ud}, EuroParl \cite{koehn2004europarl} (automatically annotated)  & 46\%\\ \hline

Mareček and Rosa 2019 \cite{marecekrosa2019balustrades} & NMT Encoder \newline (6 layers 16 heads) & Constituency & Tree induction & EuroParl \cite{koehn2004europarl} (automatically annotated) & 19\% - 33\% \\ \hline

Kim et al. 2019 \cite{kim_are_2020} & LMs (BERT, GPT2, \newline RoBERTa, XLNet) & Constituency & Tree induction & WSJ Penn Treebank \cite{marcus1993penn}, MNLI \cite{williams2018broad} & --- \\ \hline

\end{tabular}
    \caption{Summary of syntactic properties observed in Transformer's self-attention heads}
    \label{tab:syntax-in-attentions}
\end{table*}
\end{savenotes}

\begin{figure*}[t]
    \centering
    \resizebox{1.15\linewidth}{!}{\input{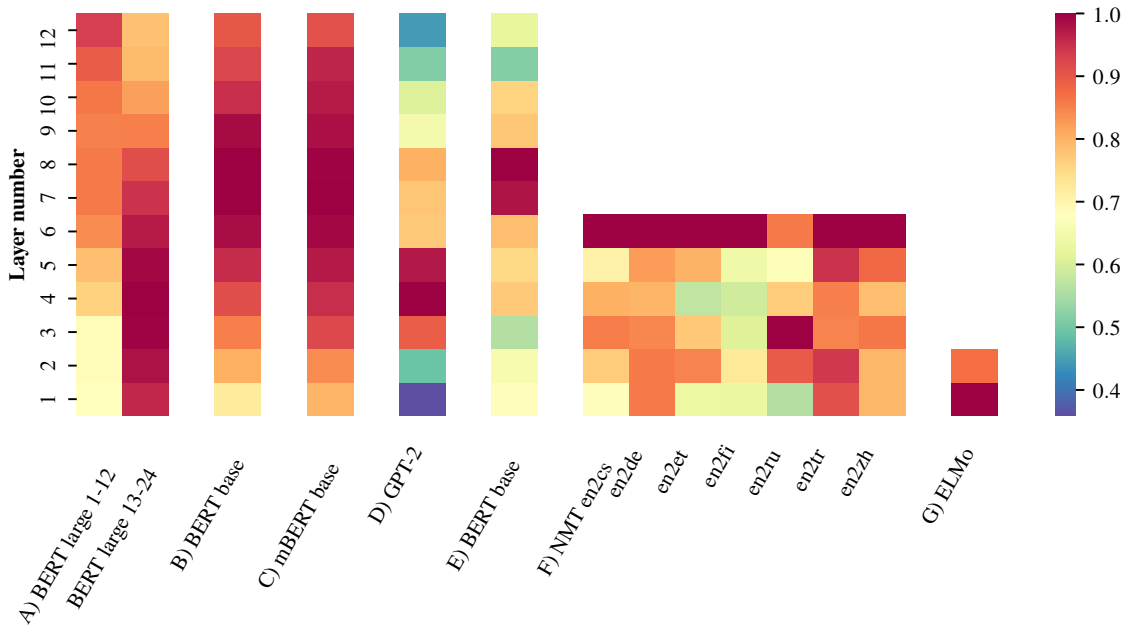}}
    \caption{Relative syntactic information across attention models and layers. The values are normalized so that the best layer for each method has 1.0.
    The methods A), B), C), and G) show undirected UAS trees extracted by probing the n-th layer \cite{hewitt-manning-2019-structural, chi2020finding}. The method D) shows the dependency alignment averaged across all heads in each layer \cite{vigbelinkov2019analyzing}. The methods E) and F) show UAS of trees induced from attention heads by the maximum spanning tree algorithm \cite{raganatotiedemann2018analysis, limisiewicz2020universal}. The results for the best layer (corresponding to value 1.0 in the plot) are: A) 82.5; B) 79.8; C) 80.1; D) 22.3; E) 24.3; F) en2cs: 23.9, en2de: 20.9, en2et: 22.1, en2fi: 24.0, en2ru: 22.4, en2tr: 17.5, en2zh: 21.6; G) 77.0}
    \label{fig:syntactic-layers}
\end{figure*}

Raganato and Tiedemann \cite{raganatotiedemann2018analysis} induce dependency trees from self-attention matrices of a neural machine translation encoder. They use the maximum spanning tree algorithm to connect pairs of tokens with high attention. 
Gold root information is used to find the direction of the edges. Trees extracted in this way are generally worse than the right-branching baseline (35.08 \% UAS on PUD) and outperform it slightly in a few heads. The maximum UAS is obtained when a dependency structure is induced from one head of the 5th layer of English to Chinese encoder - 38.87\% UAS.
Nevertheless, their approach assumes that the whole syntactic tree may be induced from just one attention head. 

Recent articles focused on the analysis of features and classification of Transformer's self-attention heads. Vig and Belinkov~\cite{vigbelinkov2019analyzing} apply multiple metrics to examine properties of attention matrices computed in a unidirectional language model (GPT-2~\cite{radford2019language}). They showed that in some heads, the attentions concentrate on tokens representing specific POS tags and the pairs of tokens are more often attended one to another if an edge in the dependency tree connects them, i.e., dependency alignment is high. They observe that the strongest dependency alignment occurs in the middle layers of the model -- 4th and 5th. They also point that different dependency types (labels) are captured in different places of the model. Attention in upper layers aligns more with subject relations whereas in the lower layer with modifying relations, such as auxiliaries, determiners, conjunctions, and expletives.

\footnotetext{A head is syntactic when the tree extracted from it surpasses the right-branching chain in terms of UAS. It is a strong baseline for syntactic trees in English. Thus only a few heads are recognized as syntactic.}
\addtocounter{footnote}{0}

Voita et al.~\cite{voita2019analyzing} also observed alignment with dependency relations in the encoders of neural machine translation systems from English to Russian, German, or French. They have evaluated dependency accuracy for four dependency labels: noun subject, direct object, adjective modifier, and adverbial modifier.
They separately address the cases where a verb attends to a dependent subject, and subject attends to governor verb. The heads with more than 10\% improvement over a positional baseline are identified as syntactic \footnote{In the positional baseline, the most frequent offset is added to the index of relation's dependent/governor to find its governor/dependent, e.g., for adjective to noun relations the most frequent offset is +1 in English}. Such heads are found in all encoder layers except the first one. In further experiments, the authors propose the algorithm to prune the heads from the model with a minimal decrease in translation performance. During pruning, the share of syntactic heads rises from 17\% in the original model to 40\% when 75\% heads are cut out, while a change in translation score is negligible. These results support the claim that the model's ability to capture syntax is essential to its performance in non-syntactic tasks.

A similar evaluation of dependency accuracy for the BERT language model was conducted by Clark et al.~\cite{clark2019does}. They identify syntactic heads that significantly outperform positional baseline for the following labels: prepositional object, determiner, direct object, possession modifier, auxiliary passive, clausal component, marker, phrasal verb particle. The syntactic heads are found in the middle layers (4th to 8th). However, there is no single head that would capture the information for all the relations.

In another experiment, Clark et al.~\cite{clark2019does} induce a dependency tree from attentions. Instead of extracting structure from each head \cite{raganatotiedemann2018analysis} they use probing to find the weighted average of all heads. The maximum spanning tree algorithm is used to induce the dependency structure from the average. This approach produces trees with 61\% UAS and can be improved to 77\% by making weights dependent on the static word representation (fixed GloVe vectors). Both the numbers are significantly higher than right branching baseline 27\%.

A related analysis for English (BERT) and the multilingual variant (mBERT) was conducted by Limisiewicz et al. \cite{limisiewicz2020universal}. 
We have observed that the information about one dependency type is split across many self-attention heads and in other cases, the opposite happens - many heads have the same syntactic function.
They extract labeled dependency trees from the averaged heads and achieves 52\% UAS and show that in the multilingual model (mBERT) specific relation (noun subject, determines) are found in the same heads across typologically similar languages.

\subsection{Constituency trees}

There are fewer papers devoted to deriving constituency syntax tree structures. 

Mare\v{c}ek and Rosa~\cite{marecekrosa2019balustrades} examined the encoder of the machine translation system for translation between English, French, and German. We observed that in some heads, stretches of words attend to the same token forming shapes similar to balustrades (Figure~\ref{fig:balusters}). Furthermore, those stretches usually overlap with syntactic phrases. This notion is employed in the new method for constituency tree induction. In their algorithm, the weights for each stretch of tokens are computed by summing the attention focused on the balustrades and then inducing a constituency tree with CKY algorithm~\cite{ney1991dynamic}. As a result, we produce trees that achieve up to 32.8\% F1 score for English sentences, 43.6\% for German and 44.2\% for French. \footnote{The evaluation was done on 1000 sentences for each language parsed with supervised Stanford Parsed} The results can be improved by selecting syntactic heads and using only them in the algorithm. This approach requires a sample of 100 annotated sentences for head selection and raises F1 by up to 8.10 percent points in English.



The extraction of constituency trees from language models was described by Kim et al.~\cite{kim_are_2020}. They present a comprehensive study that covers nine types of pre-trained networks: BERT (base, large), GPT-2 \cite{radford2019language} (original, medium), RoBERTa \cite{liu2019roberta} (base, large), XLNet \cite{yang2019xlnet} (base, large). Their approach is based on computing distance between each pair of subsequent words. In each step, they are branching the tree in the place where the distance is the highest. The authors try three distance measures on the vector outputs of the encoder layer (cosine, L1, and L2 distances for pairs of vectors) and two distance measures on the distributions of token's attention (Jason-Shannon and Hellinger distances for pairs of distribution). In the former case, distances are computed only per layer and in the latter case for each head and average of heads in one layer. The best setting achieves 40.1\% F1 score on WSJ Penn Treebank. It uses XLNet-base and Helinger distance on averaged attentions in the 7th layer. Generally, attention distribution distances perform better than vector ones. Authors also observe that models trained on regular language modeling objective (i.e., next word prediction in GPT, XLNet) captured syntax better than masked language models (BERT, RoBERTa). In line with the previous research, the middle layers tend to be more syntactic.

\subsection{Syntactic information across layers}

Figure~\ref{fig:syntactic-layers} summarizes the evaluation of syntactic information across layers for different approaches. In Transformer-based language models: BERT, mBERT, and GPT-2, the middle layers are the most syntactic. In neural machine translation models, the top layers of the encoder are the most syntactic. However, it is important to note that the NMT Transformer encoder is only the first half of the whole translation architecture, and therefore the most syntactic layers are, in fact, in the middle of the process. In RNN language model (ELMo) the first layer is more syntactic than the second one.

We conjecture that the initial Transformer's layers capture simple relations (e.g., attending to next or previous tokens) and the last layers mostly capture task-specific information. Therefore, they are less syntactic.

We also observe that in supervised probing \cite{hewitt-manning-2019-structural, chi2020finding}, better results are obtained from initial and top layers than in unsupervised structure induction \cite{raganatotiedemann2018analysis, limisiewicz2020universal}, i.e., the distribution across layers is smoother.






\section{Conclusion}
\label{sec:conclusion}




In this overview, we survey that syntactic structures are latently learned by the neural models for natural language processing tasks.
We have compared multiple approaches of others and described the features that affect the ability to capture the syntax. The following aspects tend to improve the performance on syntactic tasks such as POS tagging:
\begin{compactenum}
    \item Using contextual embeddings from RNNs or Transformer outperforms static word embeddings (Word2Vec, GloVe). 
    \item Pretraining on tasks with masked input (language modeling or machine translation) produces better syntactic representation than auto encoding.
    \item The advantage of language modeling over machine translation is the fact that larger corpora are available for pretraining.
\end{compactenum}
Our meta-analysis of latent states showed that the most syntactic representation could be found in the middle layers of the model. They tend to capture more complex relations than initial layers, and the representations are less dependent on the pretraining objectives than in the top layers. 

We have shown to what extent systems trained for a non-syntactic task can learn grammatical structures. The question we leave for further research is whether providing explicit syntactic information to the model can improve its performance on other NLP tasks.

\section*{Acknowledgments}
This work has been supported by the grant 18-02196S of the Czech Science Foundation. It has been using language resources and tools developed, stored and distributed by  theLINDAT/CLARIAH-CZ project of the Ministry of Education, Youth and Sports of the Czech Republic (project LM2018101).

\bibliography{references}

\begin{thebibliography}{10}

\bibitem{almeida2019word}
Felipe Almeida and Geraldo Xexéo.
\newblock Word embeddings: A survey.
\newblock {\em CoRR}, abs/1901.09069, 2019.

\bibitem{bahdanau2014neural}
Dzmitry Bahdanau, Kyunghyun Cho, and Yoshua Bengio.
\newblock Neural machine translation by jointly learning to align and
  translate.
\newblock {\em CoRR}, abs/1409.0473, 2015.

\bibitem{belinkov2017neural}
Yonatan Belinkov, Nadir Durrani, Fahim Dalvi, Hassan Sajjad, and James Glass.
\newblock What do neural machine translation models learn about morphology?
\newblock In {\em Proceedings of the 55th Annual Meeting of the Association for
  Computational Linguistics (Volume 1: Long Papers)}, pages 861--872,
  Vancouver, Canada, July 2017. Association for Computational Linguistics.

\bibitem{belinkov2017evaluating}
Yonatan Belinkov, Llu{\'\i}s M{\`a}rquez, Hassan Sajjad, Nadir Durrani, Fahim
  Dalvi, and James Glass.
\newblock Evaluating layers of representation in neural machine translation on
  part-of-speech and semantic tagging tasks.
\newblock In {\em Proceedings of the Eighth International Joint Conference on
  Natural Language Processing (Volume 1: Long Papers)}, pages 1--10, Taipei,
  Taiwan, November 2017. Asian Federation of Natural Language Processing.

\bibitem{blevins2018deep}
Terra Blevins, Omer Levy, and Luke Zettlemoyer.
\newblock Deep {RNN}s encode soft hierarchical syntax.
\newblock In {\em Proceedings of the 56th Annual Meeting of the Association for
  Computational Linguistics (Volume 2: Short Papers)}, pages 14--19, Melbourne,
  Australia, July 2018. Association for Computational Linguistics.

\bibitem{chi2020finding}
Ethan~A. Chi, John Hewitt, and Christopher~D. Manning.
\newblock Finding universal grammatical relations in multilingual {BERT}.
\newblock In {\em Proceedings of the 58th Annual Meeting of the Association for
  Computational Linguistics}, pages 5564--5577, Online, July 2020. Association
  for Computational Linguistics.

\bibitem{clark2019does}
Kevin Clark, Urvashi Khandelwal, Omer Levy, and Christopher~D. Manning.
\newblock What does {BERT} look at? {An} analysis of {BERT's} attention, 2019.

\bibitem{deerwester1990indexingby}
Scott Deerwester, Susan~T. Dumais, George~W. Furnas, Thomas~K. Landauer, and
  Richard Harshman.
\newblock Indexing by latent semantic analysis.
\newblock {\em Journal of the American Society for Information Science},
  41(6):391--407, 1990.

\bibitem{devlin2019bert}
Jacob Devlin, Ming-Wei Chang, Kenton Lee, and Kristina Toutanova.
\newblock Bert: Pre-training of deep bidirectional transformers for language
  understanding.
\newblock In {\em NAACL-HLT}, 2019.

\bibitem{dozat2016deep}
Timothy Dozat and Christopher~D. Manning.
\newblock Deep biaffine attention for neural dependency parsing.
\newblock In {\em 5th International Conference on Learning Representations,
  {ICLR} 2017, Toulon, France, April 24-26, 2017, Conference Track
  Proceedings}, 2017.

\bibitem{harris54}
Zellig Harris.
\newblock Distributional structure.
\newblock {\em Word}, 10(23):146--162, 1954.

\bibitem{hewitt-manning-2019-structural}
John Hewitt and Christopher~D. Manning.
\newblock A structural probe for finding syntax in word representations.
\newblock In {\em NAACL-HLT}, 2019.

\bibitem{kim_are_2020}
Taeuk Kim, Jihun Choi, Daniel Edmiston, and Sang-goo Lee.
\newblock Are {Pre}-trained {Language} {Models} {Aware} of {Phrases}? {Simple}
  but {Strong} {Baselines} for {Grammar} {Induction}.
\newblock In {\em International Conference on Learning Representations},
  January 2020.

\bibitem{koehn2004europarl}
Philipp Koehn.
\newblock Europarl: A parallel corpus for statistical machine translation.
\newblock 5, 11 2004.

\bibitem{limisiewicz2020universal}
Tomasz Limisiewicz, Rudolf Rosa, and David Mareček.
\newblock Universal dependencies according to {BERT}: both more specific and
  more general.
\newblock {\em ArXiv}, abs/2004.14620, 2020.

\bibitem{lison2018opensubtitles}
Pierre Lison, J{\"o}rg Tiedemann, and Milen Kouylekov.
\newblock {O}pen{S}ubtitles2018: Statistical rescoring of sentence alignments
  in large, noisy parallel corpora.
\newblock In {\em Proceedings of the Eleventh International Conference on
  Language Resources and Evaluation ({LREC} 2018)}, Miyazaki, Japan, May 2018.
  European Language Resources Association (ELRA).

\bibitem{liu2019linguistic}
Nelson~F. Liu, Matt Gardner, Yonatan Belinkov, Matthew~E. Peters, and Noah~A.
  Smith.
\newblock Linguistic knowledge and transferability of contextual
  representations.
\newblock In {\em NAACL-HLT}, 2019.

\bibitem{liu2020contextual}
Qi~Liu, Matt~J. Kusner, and Phil Blunsom.
\newblock A survey on contextual embeddings.
\newblock {\em ArXiv}, abs/2003.07278, 2020.

\bibitem{liu2019roberta}
Yinhan Liu, Myle Ott, Naman Goyal, Jingfei Du, Mandar Joshi, Danqi Chen, Omer
  Levy, Mike Lewis, Luke Zettlemoyer, and Veselin Stoyanov.
\newblock Roberta: A robustly optimized bert pretraining approach.
\newblock {\em arXiv preprint arXiv:1907.11692}, 2019.

\bibitem{marcus1993penn}
Mitchell~P. Marcus, Beatrice Santorini, and Mary~Ann Marcinkiewicz.
\newblock Building a large annotated corpus of {E}nglish: The {P}enn
  {T}reebank.
\newblock {\em Computational Linguistics}, 19(2):313--330, 1993.

\bibitem{marecekrosa2019balustrades}
David Mare{\v{c}}ek and Rudolf Rosa.
\newblock From balustrades to pierre vinken: Looking for syntax in transformer
  self-attentions.
\newblock In {\em Proceedings of the 2019 ACL Workshop BlackboxNLP: Analyzing
  and Interpreting Neural Networks for NLP}, pages 263--275, Florence, Italy,
  August 2019. Association for Computational Linguistics.

\bibitem{mccann2017learned}
Bryan McCann, James Bradbury, Caiming Xiong, and Richard Socher.
\newblock Learned in translation: Contextualized word vectors.
\newblock In {\em Advances in Neural Information Processing Systems}, pages
  6297--6308, 2017.

\bibitem{mikolov2013efficient}
Tomas Mikolov, Kai Chen, Greg Corrado, and Jeffrey Dean.
\newblock Efficient estimation of word representations in vector space.
\newblock {\em CoRR}, abs/1301.3781, July 2013.

\bibitem{mikolov2013linguistic}
Tomas Mikolov, Wen-tau Yih, and Geoffrey Zweig.
\newblock Linguistic regularities in continuous space word representations.
\newblock In {\em Proceedings of the 2013 Conference of the North {A}merican
  Chapter of the Association for Computational Linguistics: Human Language
  Technologies}, pages 746--751, Atlanta, Georgia, June 2013. Association for
  Computational Linguistics.

\bibitem{musil2019examining}
Tomáš Musil.
\newblock {Examining Structure of Word Embeddings with PCA}.
\newblock In {\em Text, Speech, and Dialogue}, pages 211--223. Springer
  International Publishing, 2019.

\bibitem{ney1991dynamic}
H.~{Ney}.
\newblock Dynamic programming parsing for context-free grammars in continuous
  speech recognition.
\newblock {\em IEEE Transactions on Signal Processing}, 39(2):336--340, 1991.

\bibitem{nivre2017ud}
Joakim Nivre, {\v Z}eljko Agi{\'c}, Lars Ahrenberg, Lene Antonsen, Maria~Jesus
  Aranzabe, Masayuki Asahara, Luma Ateyah, Mohammed Attia, Aitziber Atutxa,
  Elena Badmaeva, Miguel Ballesteros, Esha Banerjee, Sebastian Bank, John
  Bauer, Kepa Bengoetxea, Riyaz~Ahmad Bhat, Eckhard Bick, Cristina Bosco, Gosse
  Bouma, Sam Bowman, Aljoscha Burchardt, Marie Candito, Gauthier Caron,
  G{\"u}l{\c s}en Cebiro{\u g}lu~Eryi{\u g}it, Giuseppe G.~A. Celano, Savas
  Cetin, Fabricio Chalub, Jinho Choi, Yongseok Cho, Silvie Cinkov{\'a}, {\c
  C}a{\u g}r{\i} {\c C}{\"o}ltekin, Miriam Connor, Marie-Catherine de~Marneffe,
  Valeria de~Paiva, Arantza Diaz~de Ilarraza, Kaja Dobrovoljc, Timothy Dozat,
  Kira Droganova, Marhaba Eli, Ali Elkahky, Toma{\v z} Erjavec, Rich{\'a}rd
  Farkas, Hector Fernandez~Alcalde, Jennifer Foster, Cl{\'a}udia Freitas,
  Katar{\'{\i}}na Gajdo{\v s}ov{\'a}, Daniel Galbraith, Marcos Garcia, Filip
  Ginter, Iakes Goenaga, Koldo Gojenola, Memduh G{\"o}k{\i}rmak, Yoav Goldberg,
  Xavier G{\'o}mez~Guinovart, Berta Gonz{\'a}les~Saavedra, Matias Grioni,
  Normunds Gr{\=u}z{\={\i}}tis, Bruno Guillaume, Nizar Habash, Jan Haji{\v c},
  Jan Haji{\v c}~jr., Linh H{\`a}~M{\~y}, Kim Harris, Dag Haug, Barbora
  Hladk{\'a}, Jaroslava Hlav{\'a}{\v c}ov{\'a}, Petter Hohle, Radu Ion, Elena
  Irimia, Anders Johannsen, Fredrik J{\o}rgensen, H{\"u}ner Ka{\c s}{\i}kara,
  Hiroshi Kanayama, Jenna Kanerva, Tolga Kayadelen, V{\'a}clava Kettnerov{\'a},
  Jesse Kirchner, Natalia Kotsyba, Simon Krek, Sookyoung Kwak, Veronika
  Laippala, Lorenzo Lambertino, Tatiana Lando, Phương L{\^e}~H{\`{\^o}}ng,
  Alessandro Lenci, Saran Lertpradit, Herman Leung, Cheuk~Ying Li, Josie Li,
  Nikola Ljube{\v s}i{\'c}, Olga Loginova, Olga Lyashevskaya, Teresa Lynn,
  Vivien Macketanz, Aibek Makazhanov, Michael Mandl, Christopher Manning, Ruli
  Manurung, C{\u a}t{\u a}lina M{\u a}r{\u a}nduc, David Mare{\v c}ek, Katrin
  Marheinecke, H{\'e}ctor Mart{\'{\i}}nez~Alonso, Andr{\'e} Martins, Jan Ma{\v
  s}ek, Yuji Matsumoto, Ryan {McDonald}, Gustavo Mendon{\c c}a, Anna
  Missil{\"a}, Verginica Mititelu, Yusuke Miyao, Simonetta Montemagni, Amir
  More, Laura Moreno~Romero, Shunsuke Mori, Bohdan Moskalevskyi, Kadri
  Muischnek, Nina Mustafina, Kaili M{\"u}{\"u}risep, Pinkey Nainwani, Anna
  Nedoluzhko, Lương Nguy{\~{\^e}}n~Th{\d i}, Huy{\`{\^e}}n Nguy{\~{\^e}}n
  Th{\d i}~Minh, Vitaly Nikolaev, Rattima Nitisaroj, Hanna Nurmi, Stina Ojala,
  Petya Osenova, Lilja {\O}vrelid, Elena Pascual, Marco Passarotti,
  Cenel-Augusto Perez, Guy Perrier, Slav Petrov, Jussi Piitulainen, Emily
  Pitler, Barbara Plank, Martin Popel, Lauma Pretkalni{\c n}a, Prokopis
  Prokopidis, Tiina Puolakainen, Sampo Pyysalo, Alexandre Rademaker, Livy Real,
  Siva Reddy, Georg Rehm, Larissa Rinaldi, Laura Rituma, Rudolf Rosa, Davide
  Rovati, Shadi Saleh, Manuela Sanguinetti, Baiba Saul{\={\i}}te, Yanin
  Sawanakunanon, Sebastian Schuster, Djam{\'e} Seddah, Wolfgang Seeker, Mojgan
  Seraji, Lena Shakurova, Mo~Shen, Atsuko Shimada, Muh Shohibussirri, Natalia
  Silveira, Maria Simi, Radu Simionescu, Katalin Simk{\'o}, M{\'a}ria {\v
  S}imkov{\'a}, Kiril Simov, Aaron Smith, Antonio Stella, Jana Strnadov{\'a},
  Alane Suhr, Umut Sulubacak, Zsolt Sz{\'a}nt{\'o}, Dima Taji, Takaaki Tanaka,
  Trond Trosterud, Anna Trukhina, Reut Tsarfaty, Francis Tyers, Sumire Uematsu,
  Zde{\v n}ka Ure{\v s}ov{\'a}, Larraitz Uria, Hans Uszkoreit, Gertjan van
  Noord, Viktor Varga, Veronika Vincze, Jonathan~North Washington, Zhuoran Yu,
  Zden{\v e}k {\v Z}abokrtsk{\'y}, Daniel Zeman, and Hanzhi Zhu.
\newblock Universal dependencies 2.0 – {CoNLL} 2017 shared task development
  and test data, 2017.
\newblock {LINDAT}/{CLARIN} digital library at the Institute of Formal and
  Applied Linguistics ({{\'U}FAL}), Faculty of Mathematics and Physics, Charles
  University.

\bibitem{pennington2014glove}
Jeffrey Pennington, Richard Socher, and Christopher~D. Manning.
\newblock Glove: Global vectors for word representation.
\newblock In {\em Empirical Methods in Natural Language Processing (EMNLP)},
  pages 1532--1543, 2014.

\bibitem{peters2018deep}
Matthew~E. Peters, Mark Neumann, Mohit Iyyer, Matt Gardner, Christopher Clark,
  Kenton Lee, and Luke Zettlemoyer.
\newblock Deep contextualized word representations.
\newblock In {\em Proceedings of the 2018 Conference of the North {A}merican
  Chapter of the Association for Computational Linguistics: Human Language
  Technologies, Volume 1 (Long Papers)}, New Orleans, Louisiana, June 2018.
  Association for Computational Linguistics.

\bibitem{radford2019language}
Alec Radford, Jeff Wu, Rewon Child, David Luan, Dario Amodei, and Ilya
  Sutskever.
\newblock Language models are unsupervised multitask learners.
\newblock 2019.

\bibitem{raganatotiedemann2018analysis}
Alessandro Raganato and J{\"o}rg Tiedemann.
\newblock An analysis of encoder representations in transformer-based machine
  translation.
\newblock In {\em Proceedings of the 2018 {EMNLP} Workshop {B}lackbox{NLP}:
  Analyzing and Interpreting Neural Networks for {NLP}}, pages 287--297,
  Brussels, Belgium, November 2018. Association for Computational Linguistics.

\bibitem{vaswani2017attention}
Ashish Vaswani, Noam Shazeer, Niki Parmar, Jakob Uszkoreit, Llion Jones,
  Aidan~N. Gomez, Lukasz Kaiser, and Illia Polosukhin.
\newblock Attention is all you need.
\newblock In {\em Advances in Neural Information Processing Systems 30: Annual
  Conference on Neural Information Processing Systems 2017, 4-9 December 2017,
  Long Beach, CA, {USA}}, pages 5998--6008, 2017.

\bibitem{vig2019transformervis}
Jesse Vig.
\newblock A multiscale visualization of attention in the transformer model.
\newblock In {\em Proceedings of the 57th Conference of the Association for
  Computational Linguistics, {ACL} 2019, Florence, Italy, July 28 - August 2,
  2019, Volume 3: System Demonstrations}, pages 37--42. Association for
  Computational Linguistics, 2019.

\bibitem{vigbelinkov2019analyzing}
Jesse Vig and Yonatan Belinkov.
\newblock {Analyzing the Structure of Attention in a Transformer Language
  Model}.
\newblock In {\em Proceedings of the 2019 ACL Workshop BlackboxNLP: Analyzing
  and Interpreting Neural Networks for NLP}, pages 63--76, Florence, Italy,
  August 2019. Association for Computational Linguistics.

\bibitem{voita2019analyzing}
Elena Voita, David Talbot, Fedor Moiseev, Rico Sennrich, and Ivan Titov.
\newblock Analyzing multi-head self-attention: Specialized heads do the heavy
  lifting, the rest can be pruned.
\newblock In {\em Proceedings of the 57th Annual Meeting of the Association for
  Computational Linguistics}, pages 5797--5808, Florence, Italy, July 2019.
  Association for Computational Linguistics.

\bibitem{Wang-2019}
Yuxuan Wang, Wanxiang Che, Jiang Guo, Yijia Liu, and Ting Liu.
\newblock Cross-lingual bert transformation for zero-shot dependency parsing.
\newblock {\em Proceedings of the 2019 Conference on Empirical Methods in
  Natural Language Processing and the 9th International Joint Conference on
  Natural Language Processing (EMNLP-IJCNLP)}, 2019.

\bibitem{williams2018broad}
Adina Williams, Nikita Nangia, and Samuel Bowman.
\newblock A broad-coverage challenge corpus for sentence understanding through
  inference.
\newblock In {\em Proceedings of the 2018 Conference of the North {A}merican
  Chapter of the Association for Computational Linguistics: Human Language
  Technologies, Volume 1 (Long Papers)}, pages 1112--1122, New Orleans,
  Louisiana, June 2018. Association for Computational Linguistics.

\bibitem{yang2019xlnet}
Zhilin Yang, Zihang Dai, Yiming Yang, Jaime~G. Carbonell, Ruslan Salakhutdinov,
  and Quoc~V. Le.
\newblock Xlnet: Generalized autoregressive pretraining for language
  understanding.
\newblock In {\em NeurIPS}, 2019.

\bibitem{zhang2018language}
Kelly~W. Zhang and Samuel~R. Bowman.
\newblock Language modeling teaches you more syntax than translation does:
  Lessons learned through auxiliary task analysis.
\newblock In {\em Proceedings of the 2018 {EMNLP} Workshop {B}lackbox{NLP}:
  Analyzing and Interpreting Neural Networks for {NLP}}, November 2018.

\end{thebibliography}

\end{document}